\documentclass [11pt,letterpaper]{article}
\usepackage{lmodern}
\usepackage{amsmath, amssymb}
\usepackage{graphicx} 
\usepackage{algorithm}
\usepackage{algpseudocode}
\usepackage{caption}
\usepackage{float}
\usepackage{url}
\usepackage{doi}
\usepackage[a4paper, top=1 in, bottom=1 in, left=1in, right=1in]{geometry}

\title{A Multi-Component Reward Function with Policy Gradient for Automated Feature Selection with Dynamic Regularization and Bias Mitigation}

\author{
    \begin{minipage}{0.48\textwidth}\centering
        Sudip Khadka\\[-0.3ex]
        \texttt{\small{skhadka3@terpmail.umd.edu}}
    \end{minipage}\hfill
    \begin{minipage}{0.48\textwidth}\centering
        L.S. Paudel\\[-0.3ex]
        \texttt{\small{laxmansharmapaudel@gmail.com}}
    \end{minipage}
}
\date{}
\begin{document}

\maketitle

\begin{abstract}

Static feature exclusion strategies often fail to prevent bias when hidden dependencies influence the model predictions. To address this issue, we explore a reinforcement learning (RL) framework that integrates bias mitigation and automated feature selection with in a single learning process.  Unlike traditional heuristic-driven filter or wrapper approaches, our RL agent adaptively selects features using a reward signal that explicitly integrates predictive performance with fairness considerations. This dynamic formulation allows the model to balance generalization, accuracy, and equity throughout the training process, rather than rely exclusively on pre-processing adjustments or post hoc correction mechanisms. In this paper, we describe the construction of a multi-component reward function, the specification of the agent’s action space over feature subsets, and the integration of this system with ensemble learning.  We aim to provide a flexible and generalizable way to select features in environments where predictors are correlated and biases can inadvertently re-emerge.
\\


\end{abstract}

\section{Introduction}
High-dimensional data can present substantial challenges in supervised machine learning through feature redundancy and the inclusion of irrelevant variables  \cite{Kanvinde2023, Ajibade2024}. These issues are widely recognized to affect the  predictive performance and generalizability of the machine learning (ML) models \cite{Ajibade2024}. Various dimensionality reduction methods, such as principal component analysis \cite{Salem2019, Ramasubramanian2024}, manifold learning techniques \cite{Altaibek2024}, and more recent sparse regularization approaches \cite{Zou2005, Allerbo1970}, each seek to overcome these challenges. Despite such progress, effective feature selection technique has remained a central area of investigation, especially in socially sensitive domains such as finance and healthcare. Standard dimensionality reduction methods do not consider the causal relationships between variables which can lead to biased or discriminatory outcomes because they are specifically designed to select features that maximize predictive performance. Another common feature selection practice involves identifying important features through permutation important and gini impurity \cite{Hapfelmeier2013}, and then excluding those biased or sensitive features based on domain knowledge. However, this approach rarely eliminates bias and in some instances it may reduce the model generalization ability or generates hidden forms of unfairness through the latent leakage \cite{Zhang2024, Zhao2021}. Post hoc interpretability frameworks including SHapley Additive Explanations (SHAP) and Local Interpretable Model-agnostic Explanations (LIME) can provide valuable insights into feature contributions \cite{hlongwane2024, Loecher1970}, but cannot directly mitigate the presence of biased variables during training.

Empirical evidence from multiple application has shown that excluding sensitive attributes such as race, gender, or age from predictive models is insufficient to prevent discriminatory outcomes \cite{Obermeyer2019, Engel2024}. The presence of correlated features and structural inequities often reintroduce bias that leads to disparate impacts between the demographic groups. For example, the Consumer Financial Protection Bureau (CFPB) reported that credit scoring models continued to exhibit racial disparities even after excluding the zip code. This phenomenon occurred because the correlated variables, such as income and type of employment, led to reintroduction bias through latent leakage. This concern is presented in \cite{Fuster2022}, which revealed that ML in credit scoring can intensify racial disparities due to increased model flexibility and their ability to capture complex relationships in the data. Socioeconomic variables such as income and type of employment can act as proxies for sensitive attributes, thereby amplifying potential discriminatory effects. Regulatory guidance from the Bank for International Settlements \cite{BIS2025} further underscores the importance to address such issues by focusing on a fairness aware and transparent model to avoid discriminatory practices in credit lending. A similar phenomenon has been documented in healthcare, where a widely deployed risk prediction algorithm underestimated the care needs of black patients \cite{Obermeyer2019}. Although race was not explicitly included as a predictor, the model relied on healthcare costs as a proxy for health care needs. This systematically disadvantaged black patients who historically suffered lower medical expenditures due to access barriers. These limitations highlight the need for  fairness aware feature selection approaches that remain robust in both high-dimensional and correlated predictor settings.

 As a result of these challenges, Reinforcement Learning (RL) has recently emerged as a promising paradigm for feature selection for structural datasets. Stochastic and probabilistic decision making frameworks for feature selection frame feature selection problem as a sequential decision making task\cite{sutton2020, Shaer2024}. These framework include approaches such as  dynamic feature selection and in-training feature masking, which treat feature selection as an adaptive procedure rather than a fixed preprocessing step \cite{Shaer2024, Fan2022, Li2022,  JIAJING2023}. Several studies have confirmed that RL agents can dynamically identify and prioritize relevant predictors during training that result in improved accuracy and robustness of the model \cite{Shaer2024, Fan2022, Li2022, Kim2024, Nagaraju2025, Bouchlaghem2024}. These approaches provide a data-driven and adaptive alternative to conventional static feature selection heuristics. 
 A RL method that employed Monte Carlo sampling was proposed to efficiently explore feature subsets to optimize classification performance and  minimized computational overhead\cite{Liu2021}. In a similar manner \cite{Kim2024} proposed, a multi-agent RL that used primary reward signal to enable the identification of the most informative features and improved classification performance. These studies show that RL can navigate complex combinatorial feature spaces more effectively than traditional greedy approaches. Their focus remains mainly on optimizing predictive performance, without explicitly addressing fairness or accounting for causal relationships among features. To address these concerns, \cite{Ling2023} adopted causal graphs method to capture both direct and indirect effects of sensitive attributes by selecting features that reduces bias while maintaining predictive accuracy. Likewise, \cite{Belitz2021} proposed a procedural fairness methods to exclude features from being used in model training that could introduce discriminatory effects. It imposed an unfairness weight to penalize features that contribute to group based disparities during model selection.  Although these methods successfully improve fairness, they do not incorporate reinforcement learning techniques nor integrate performance-driven rewards with fairness penalties within a single unified framework. These limitations underscore the need for approaches that jointly consider dimensionality reduction and fairness constraints within the same learning process. 
 
 To address these challenges, we proposed a multi component reward functions in the RL learning dynamic to balance compactness with equitable treatment across groups. We used the decision tree as a black-box learner and introduce a policy that regulates feature influence through sequential interaction. The reward function directs the policy towards its intended objective by incorporating performance, fairness and compactness. The Performance component measures the classification performance of the model using Area Under the Curve (AUC) score. The fairness components consist of direct and indirect penalties that discourages the policy to select biased features or those casually linked to them. Finally, the compactness include size penalty and shaped reward to encourage parsimonious feature subsets that retain predictive utility while reducing redundany.  These reward-based penalties operate as soft constraints that guide the agent's behavior toward actions that maximize expected returns and simultaneously avoid the reliance on sensitive or biased predictors.  We applied this model to a credit risk dataset to predict loan default, and it outperformed all baseline models when benchmarked using statistical metrics such as Receiver operating characteristic (ROC) curve and bias score plots. To assess policy convergence, we examined model learning behaviour using a  state-action value heatmap, performance scatterplot, penalty trajectory, and reward trajectory, all of which demonstrated stable convergence. This approach offers a principled mechanism for implementing RL in high-stakes, fairness-sensitive domains where biased features can lead to systemic social consequences.

\section{Proposed Model}
\subsection{Agent Architecture and Learning Dynamic Overview}
 We begin by describing the design of our reward architecture, and then explained how we integrate it into a policy gradient training framework. Our model is based on the classic Markov Decision Process (MDP) \cite {sutton2020}, but augments the reward signal to better capture bias dynamics. We assign a reward \(R_t = r\)  based on the immediate outcome of action \(A_t = a\) in the environment state \(S_t\), combining performance-based rewards with penalty components designed to discourage biased behavior. These penalties are embedded directly into the reward function, which enables the agent to balance task performance with fairness constraints.

\begin{figure}[H]
    \centering
    \includegraphics[width=0.65\textwidth]{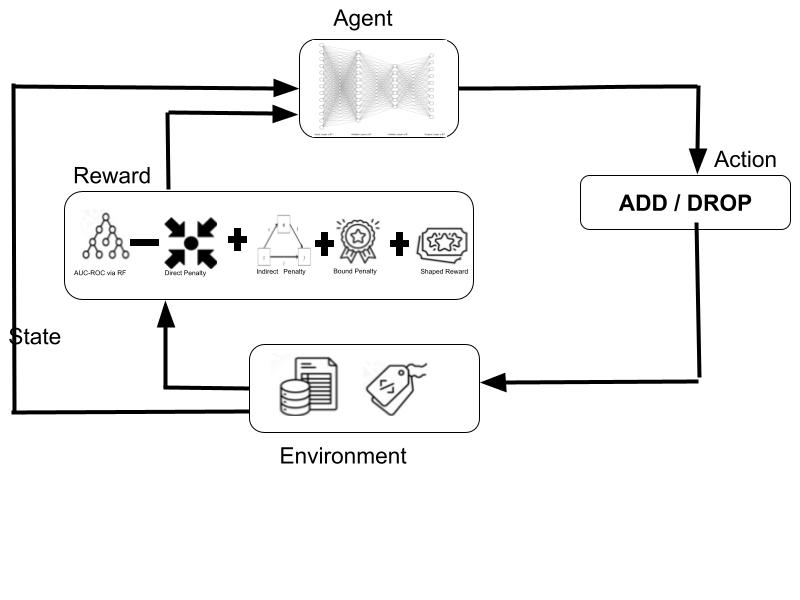}
    \vspace{-45pt}
    \caption{The model architecture consists of the standard reinforcement learning components: agent, actions, environment, state, and reward. The agent is parameterized as a multi-layer perceptron (MLP) with ReLU activations and a softmax output layer to approximate the action probability distribution.}     
\end{figure}

\

In this formulation, the agent is governed by the policy \(\pi\), and serves as a feature selection system that operates within the feature space (i.e. environment) \cite{sutton2020}. Each \(s \in S\) represents a specific subset of selected features at a given timestep, and an action \(a \in A\) corresponds to either inclusion or exclusion of the candidate feature from the active subset\cite{sutton2020}. Following the policy, the environment transitions form the current state \(s\) to the next state \(s'\) by following an action \(a\), and returns a scaler reward \(R(s,a)\) \cite{sutton2020, Shaer2024}.  The reward reflects the predictive performance of a base model trained on the resulting feature subset, adjusted for penalties and potential rewards.

The policy optimization is expressed as maximizing the expected discounted return under \(\pi_\theta\). Given a trajectory \(\tau=(S_0,a_0,..., s_{T-1}, a_{T-1}\)) generated by \(\pi_\theta\), the objective is defined as \cite{sutton2020,Shaer2024};

\[
J(\pi_\theta) = \mathbb{E}_{\tau \sim \pi_{\theta}}\left[\sum_{t=0}^{T-1}R(s_t, a_t, s\prime)\right]
\]

Unlike constrained Markov Decision Process (CMDP) \cite{JIAJING2023, Achiam2013,  Chow2018} approaches, we do not directly impose explicit optimization constraints. Instead, we encode penalties and incentives directly into the reward function. This approach avoids the complexity of handling constraints explicitly, making implementation simpler while still guiding the agent towards the intended behavior. 

We define reward to reflect the policy's predictive accuracy that is evaluated under a AUC score of the generic model. To mitigate the undesired influence of biased features, we supplement the reward with two penalty terms. The direct penalty which is applied when a known biased feature is chosen, and this is a fixed cost based on domain knowledge. On the other hand, an indirect penalty is estimated from the inferred causal links between biased and selected features. To capture this casual relationship between the variables, we build a graph whose nodes are chosen features and the edges capture strong interference correlations, and derive a penalty from the strength of these inferred paths. This mechanism is adapted from graph-based regularization approaches in feature selection \cite{Giarelis2020, Xie2022, Qian2021}.  We extended the mechanism in our work to approximate causal leakage rather than pure correlation. To encourage a parsimonious feature subset aligned with domain relevance, we also regularize for sparsity and impose soft constraints. If the selected set is too small or too large compared to the preset bounds, we apply a further penalty. Together, these design choices ensure that our reward function balances predictive accuracy with fairness and sparsity. This results in a feature set that is both relevant and well-constrained. Formally, this can be expressed as;

{\small
\[
\begin{split}
R(s_t,a_t)&=\underbrace{W}_{AUC \ Score}-\underbrace{\sum_{f\in S\cap B}\psi}_{\text{direct penalty}}
+\underbrace{\sum_{f_b\in B\cap V(G)}\sum_{\substack{f_s\in S\cap V(G)\\ f_s\neq f_b}}
\mathbf{1}\{P(f_b,f_s)\}\frac{w(f_b,f_s)\,\lambda}{\ell(f_b,f_s)}}_{\text{indirect penalty}}\\
&\quad+\underbrace{\phi(|S|)}_{\text{size penalty}}+\underbrace{\sum_{f\in S\cap R}\rho}_{\text{shaped reward}}.
\end{split}
\]
}

\(W\) represents the predicted utility of the selected feature set and is estimated using a baseline model to predict the quality of feature subsets based on historical data. It serves as a baseline reward that captures the inherent value of features before accounting for penalties or additional constraints. \(S\) denotes the set of selected features, while \(B\) represents the set of biased or sensitive features that may introduce undesirable effects. We modeled the indirect penalty using a graph \cite{Giarelis2020} \(G\) where the nodes \(V(G)\) correspond to all the features under consideration.  We define an indicator function \(\mathbf{1}\{P(f_b, f_s)\}\) that evaluates to \(1\) is there is a path between a biased feature \(f_b \in B\) and a selected feature \(f_s \in S \in G\), and \(0\) otherwise, allowing the reward to account for the indirect influence between features. Each path is associated with a weight \(w(f_b, f_s)\) which corresponds to the Pearson correlation between features \cite{Xie2022}, and a length \(l(f_b, f_s)\) represents the Euclidean distance between two points \cite{Qian2021}. Together with the indirect factor \(\lambda\), these elements determine the magnitude of the indirect penalty. The direct cost of selecting a biased feature is given by \(\psi\), and additional positive reward contributions for desired features are captured by \(\rho\). To regulate the number of selected features, the size penalty \(\phi(|S|)\), is defined as

\vspace{12pt}
\[
    \phi(|S|) = 
    \begin{cases} 
      (M_{\min} - |S|)\cdot \xi, & |S| < M_{\min}, \\[6pt]
      (|S| - M_{\max})\cdot \xi, & |S| > M_{\max}, \\[6pt]
      0, & \text{otherwise},
    \end{cases}
\]
\vspace{12pt}

Where \(M_{min}\) and \(M_{max}\) represent the minimum and maximum allowable number of features, respectively. \(\xi\) is the constant weighting factor. This formulation ensures that the reward function balances the utility of selected features with both direct and indirect penalties, while also enforcing constraints on feature set size. 

 We optimize the policy using the REINFORCE algorithm \cite{Williams1992, Sutton2018}, as its stochastic policy-gradient framework naturally balances exploration and exploitation, allowing the model to discover informative feature subsets without relying on a greedy search method. The gradient of the expected return is given by \cite{Williams1992, Sutton2018}
 \[
\nabla J(\pi_{\theta})
= \mathbb{E}_{\tau \sim \pi_{\theta}}
\left[
\sum_{t=0}^{T-1}
\nabla_{\theta} \log \pi_{\theta}(a_t \mid s_t)\, G_t
\right].
\]

For each episode, we accumulate the returns \(G_t = R(s_t, a_t)\) and perform a gradient ascent on the policy parameters \cite{Williams1992, Sutton2018}. 
 $$\theta \leftarrow \theta + \alpha \sum_{t=0}^{T-1} G_t \nabla \log \pi_\theta(A_t | S_t)$$

This update rule corresponds to the REINFORCE algorithm, where the gradient is estimated using the observed return from each episode. In this formulation, \(\alpha\) is the learning rate, which controls the step size during parameter updates, while \(T\) represents the episode length that defines the time horizon over which the trajectory is evaluated \cite{Williams1992, Sutton2018}. At each discrete time step \(t\), the agent transitions from state \(S_t\) to action \(A_t\) following a stochastic policy \(\pi_\theta\) \cite{Williams1992, Sutton2018}. The term \(\nabla \ log \ \pi_\theta (A_t|S_t)\) refers to the gradient of log-probability of selecting action \(A_t\) in state \(S_t\) under the policy \(\pi_\theta\), and plays a central role in updating the policy parameters \cite{Williams1992, Sutton2018}. The policy optimization is guided by a multi-component reward function that jointly captures model accuracy, fairness, and efficiency. This reward structure enables the agent to learn a strategy that balances these objectives while maximizing cumulative rewards.

\subsection{Model Integration and Training}
 To operationalize this framework in a practical learning context, we adopted a standard Decision Tree Classifier as a predictor, which returns the output AUC score, initialized as  \(W\).  Given their proven effectiveness in classification tasks on tabular data- especially in credit risk modeling \cite{Golbayani2020, Rada2017, Yan2022, Han2024} - we selected the Decision Tree as the generic model for this work. The selected features are subsequently used to train a decision tree classifier that serves as the base predictive model. The decision tree is not part of the differentiable computation graph, as the policy network does not backpropagate through it \cite{Vos2024}. Instead, the tree serves exclusively as a black-box reward evaluator within the learning framework. The goal of our policy is to learn a feature selection strategy \(\pi_\theta(a|x)\) that incrementally constructs a subset of features through sequential actions. At each step \(t\), the state \(s_t\) encodes the current feature subset, and the action \(a_t\) corresponds to adding or removing a feature. The input is then masked as \(\tilde{x}= x^{(S_t)}\), and a Random forest classifier  is trained on the masked training set and evaluated on the validation set. The AUC score at step \(t\) is defined as the Random Forest validation performance, while the corresponding reward is a penalty-adjusted reward at time \(t\). The algorithm for implementing the proposed model is summarized as follows:

\renewcommand{\thealgorithm}{}
\begin{algorithm}[H]
\caption{: Reinforcement learning algorithm for fair feature selection}
\begin{algorithmic}
\State \textbf{Define environment:} Create a FeatureSelectionEnv class:
    \State \hspace{1.2cm} Initialize with all available features
    \State \hspace{1.2cm} Maintain the current set of selected features (state)
    \State \hspace{1.2cm} Implement a step function to add or remove a feature based on action

\State \textbf{Define reward function:} 
    Input: current set of selected features
    \Statex \hspace{1.2cm} If no features selected $\rightarrow$ reward = 0
    \Statex \hspace{1.2cm} Train RandomForestClassifier and compute AUC Score
    \Statex \hspace{1.2cm} Apply penalties: 
        \Statex \hspace{2.3cm}Direct (biased features), 
        \Statex \hspace{2.3cm}Indirect (causal graph correlations), 
    \Statex \hspace{1.2cm} Add constraint (feature count bounds)
    \Statex \hspace{1.2cm} Add rewards for predefined features
    \Statex \hspace{1.2cm} Return total reward = ROC AUC $-$ penalties $+$ feature rewards

\State \textbf{Define policy network:}
    Create \text{PolicyNetwork} (PyTorch)
    \Statex \hspace{1.2cm} Input: Binary vector of feature selections
    \Statex \hspace{1.2cm} Output: $2 \times$ (no. features) $\rightarrow$ add/remove actions
    \Statex \hspace{1.2cm} Layers: ReLU activations, Softmax for action probabilities

\State \textbf{Define REINFORCE agent:}
    Create \text{REINFORCE} class with policy 
    \State \hspace{4.9cm} network, learning rate, discount rate
    \Statex \hspace{1.2cm} \text{Select\_action:}
        Input state $\rightarrow$ policy $\rightarrow$ sample action, store log-prob, 
         \State \hspace{3.4cm}Return action
    \Statex \hspace{1.2cm} \text{Update\_policy:}
        Compute discounted returns, policy loss, backprop 
        \State \hspace{3.4cm}via optimizer, retset eposide memory
    \Statex \hspace{1.2cm} \text{Store\_reward:} Record each step

\State \textbf{Training loop:}
    \Statex \hspace{1cm} - Initialize environment, policy network, and REINFORCE 
    \Statex \hspace{1.4cm}agent
    \Statex \hspace{1cm} - Define training parameters (episodes, steps)
    \Statex \hspace{1cm} - Track best reward, accuracy, and selected features
    \Statex \hspace{1cm} - For each eposide:
        \Statex \hspace{2.4cm} Reset environment, initialize reward
        \Statex \hspace{2.4cm} For each step:
            \Statex \hspace{3cm}Get state $\rightarrow$ tensorize $\rightarrow$ select action
            $\rightarrow$ map action $\rightarrow$ 
            \Statex \hspace{3cm}environment step
            $\rightarrow$ Caluclate reward $\rightarrow$ store reward
            $\rightarrow$ 
            \Statex \hspace{3cm}Accumulate states (rewars, accuracy)
        \Statex \hspace{1cm} - Update policy; print eposide statistics
        \Statex \hspace{1cm} - Update best-performing eposide
    
\State \textbf{Output results:}
    Print the best total reward, best AUC score, and best selected 
    \Statex \hspace{2.9cm}features. Optionally, plot the causal graph for the  best selected features

\end{algorithmic}
\end{algorithm}

\section{Practical Implementation}
We chose credit scoring as the application context to evaluate the proposed model because this domain is highly sensitive to bias and involves complex, high-dimensional data. It offers a practical and rigorous setting to test how well feature selection and dimensionality reduction methods perform in real-world, high-stakes situations. We used the Credit Default Risk dataset \cite{Anna2018} from Kaggle, which contains 742,935 observations across four loan types: consumer credit, credit card, car loan, and microloan. The dataset includes 158 attributes, with a mix of categorical and continuous variables. We handled missing values in the exogenous variables by replacing them with a designated sentinel value. We predefined biased features—such as age, gender, and marital status—based on the protected classes identified by the CFPB under the Equal Credit Opportunity Act \cite{BIS2025, MRM2011}. To guide feature selection, we set a lower limit of 8 and an upper limit of 20 features, allowing the model to identify the most informative subset. We set additional reward constrain to prioritized key financial indicators, including total income, credit amount, and annuity amount.

The features selected by the proposed model are not arbitrary choices aimed solely at enhancing predictive accuracy; rather, they are identified through rigorous statistical testing and validation as both statistically and practically relevant for credit risk prediction. Among the 158 candidate variables, eight features—including \texttt{ MONTHS BALANCE, CREDIT CURRENCY, AMT CREDIT, AMT INCOME TOTAL, REGION RATING CLIENT, LIVE CITY NOT WORK CITY, and} \\ \texttt{CNT DRAWINGS POS CURRENT} — were selected based on their demonstrated predictive power. These selections are consistent with regulatory and industry best practices and effectively capture critical facets of client behavior, financial exposure, and regional risk \cite{BIS2025, MRM2011}. This approach ensures that the model’s feature selection process is meaningful, transparent, fair, and robust, thus providing a principled foundation for automated feature selection in credit default prediction.

\begin{figure}[H]
    \centering
    \includegraphics[width=0.6\textwidth]{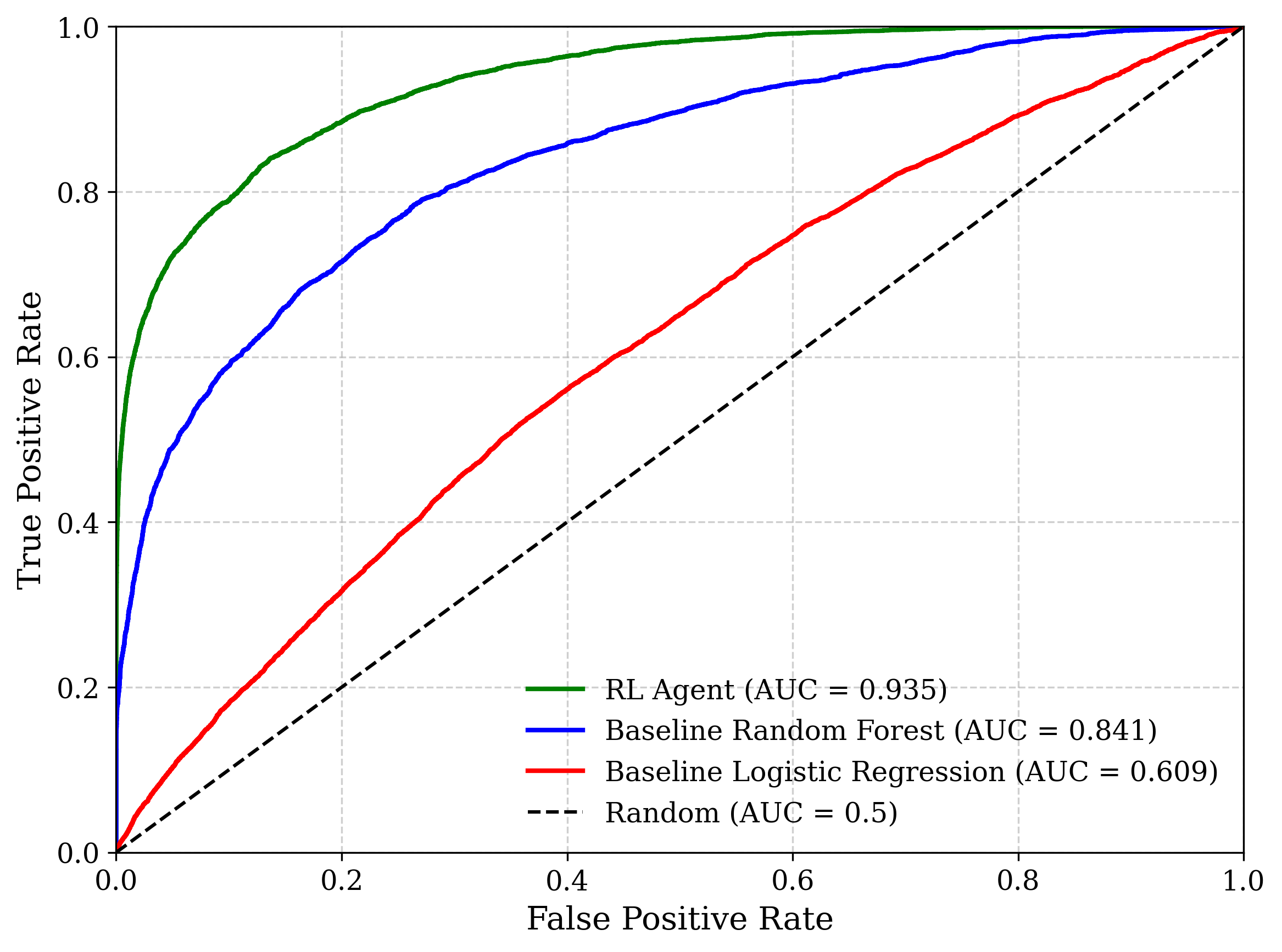}
    \caption{ROC curve comparing the performance of the RL policy with feature selection against baseline random forest and logistic regression models. This figure illustrates that our proposed model outperforms the baseline random forest and logistic regression models. }
    \label{fig:roc}
\end{figure}

We conducted an empirical analysis of the proposed approach in our datasets. The results, shown in Fig. \ref{fig:roc}, compare our model with established baseline methods. Our model outperformed both logistic regression and random forest on the same dataset. This outcome demonstrates the model’s ability to identify the most effective combination of features that enhances out-of-sample predictive performance.

The predictive accuracy we used is the AUC score, and the bias score is the sum of both direct and indirect bias components. To quantify the contribution of each feature to potential bias, we construct a correlation-based graph \(G=(V,E)\) where each node \(v_i \in V\) represents a feature. An edge \(v_i, v_j \in E\) is added between features \(i\) and \(j\) if there exists an absolute correlation. Each edge is assigned a weight \(w_{ij}\) that corresponds to a correlated value as described in \cite{Giarelis2020}. We assign a fixed score (8 in this case) to features that belong to the direct bias feature. For the indirect feature, we assign a score based on their proximity in the graph. A score of zero is assigned to features with no connection to biased features. Finally, we compute the total bias score for a set of features by summing the individual bias scores. The score for feature \(v_i\) is defined as:

\[
P(v_i) =
\begin{cases}
\psi , & \text{if } v_i \in B, \\[1mm]
\dfrac{\lambda}{d_E(v_i, B)}, & \text{if } v_i \notin B \text{ and } \exists b \in B \text{ such that a path exists}, \\[1mm]
0, & \text{if } v_i \notin B \text{ and } \forall b \in B, \text{ no path exists},
\end{cases}
\]

where $\lambda$ is a scaling constant (e.g., $k = 3$), $\psi$ is a constant direct penalty (e.g., $\psi = 8$), and
\[
d_E(v_i, B) = \min_{b \in B} \| \mathbf{v_i} - \mathbf{b} \|_2
\]
denotes the Euclidean distance between the feature $v_i$ and the closest sensitive feature $b$.

The total bias score is then computed as:

\[
P_{\text{total}} = \sum_{v_i \in V} P(v_i)
\]

To illustrate how the proposed bias score quantifies feature biases, lets consider two cases.

\paragraph{Case 1.} 
Let the sensitive feature set be 
\( D = \{ \text{Age}, \text{Gender} \} \), and assume the model selects 
\( V = \{ \text{Age}, \text{Level of Education}, \text{Income} \} \). 
A graph based dependency analysis shows a direct link between \(\textit{Age}\) and \(\textit{Level of Education}\) with a distance of 1.5, while \(\textit{Income}\) remains disconnected from the sensitive feature. 
Applying the penalty function with \(\psi = 8\) and \(\lambda = 3\), the direct inclusion of  \(\textit{Age}\) incurs the maximum penalty 
\( P(\text{Age}) = 8 \), where as  \(\textit{Level of Education receives}\) a reduced penalty proportional to its distance from Age: 
\[
P(\text{Level of Education}) = \frac{3}{1.5} = 2.
\] 
Since Income is independent, \( P(\text{Income}) = 0 \). The total bias score for this case is 
\( P_{\text{total}} = 10 \)

\paragraph{Case 2.} 
In the second scenario, let the sensitive feature set be 
$D = \{\text{Marital Status}, \text{Race}\}$, and assume the model selects 
$V = \{\text{Race}, \text{Marital Status}, \text{Income}, \text{Credit Line}\}$.
Both \textit{Race} and \textit{Marital Status} are sensitive features, 
while \textit{Income} and \textit{Credit Line} are not connected to them 
through the graph. Applying the penalty function with $\psi = 8$ and $\lambda = 3$, 
the direct inclusion of sensitive features results in 
$P(\text{Race}) = 8$ and $P(\text{Marital Status}) = 8$. 
Since \textit{Income} and \textit{Credit Line} are independent 
of the sensitive nodes, their penalties are zero, i.e., 
$P(\text{Income}) = P(\text{Credit Line}) = 0$. 
The total bias score is therefore

\[
P_{\text{total}} = 8 + 8 + 0 + 0 = 16.0
\]

\

\begin{figure}[H]
    \centering
    \includegraphics[width=0.7\textwidth]{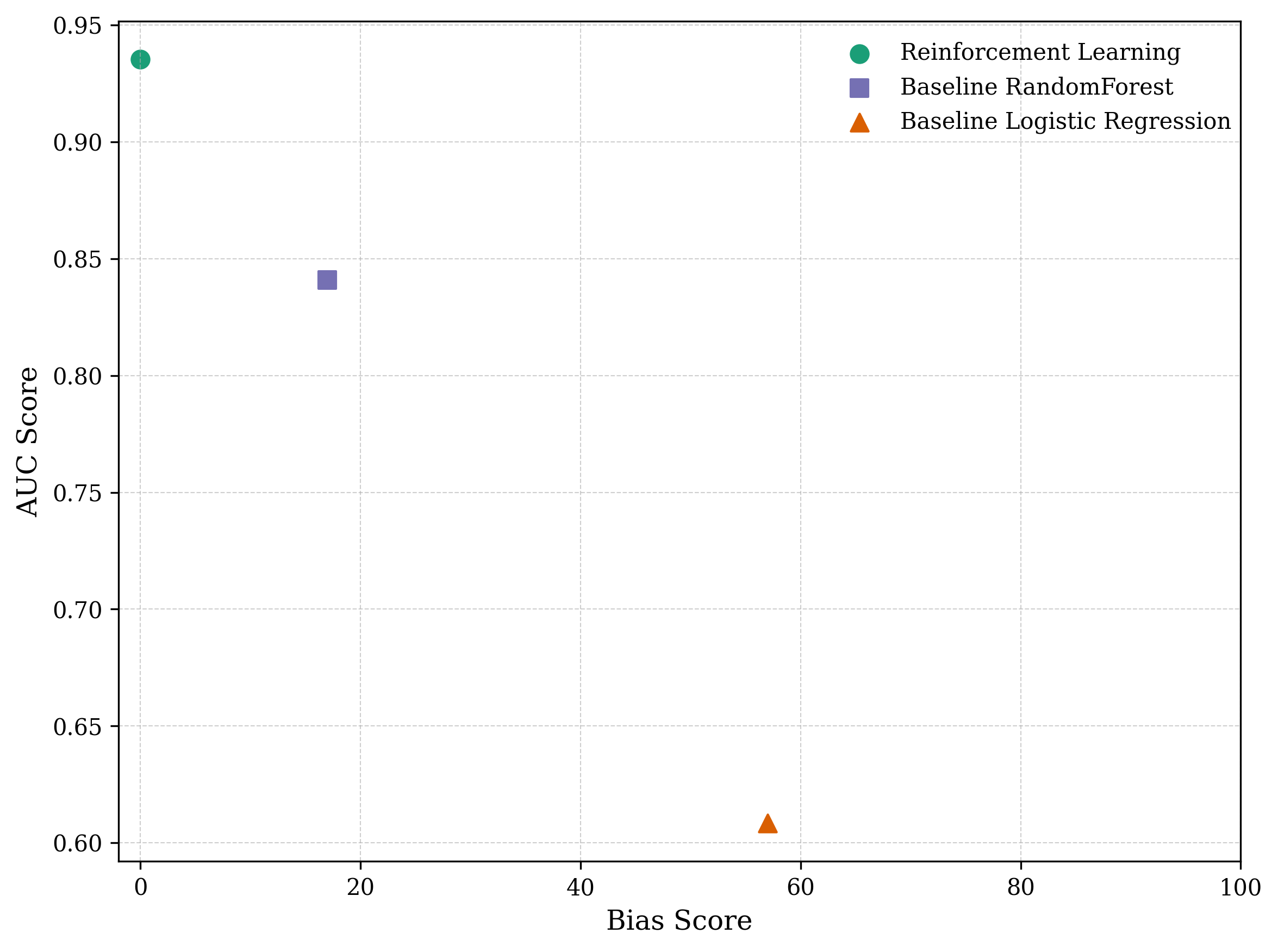}
    \caption{This figure illustrates the relationship between AUC and bias scores across different models. The comparison accounts for both direct and indirect sources of bias, providing insight into each model’s balance between predictive accuracy and fairness.}
    \label{fig:bias_accuracy}
\end{figure}

Fig. \ref{fig:bias_accuracy} depicts the trade-off relationship between bias and predictive accuracy. Our proposed RL model stands out as both the most accurate and the least biased.  The baseline random forest achieved a decent accuracy but carried more bias. The baseline Logistic Regression model performed poorly on both dimensions, with a relatively low AUC score and the highest bias score. These findings demonstrate that reducing bias does not necessarily require compromising accuracy. In this case, the RL model offers the most effective balance between the two objectives.

\begin{figure}[H]
    \centering
    \begin{minipage}[b]{0.49\textwidth}
        \centering
        \includegraphics[width=0.95\textwidth]{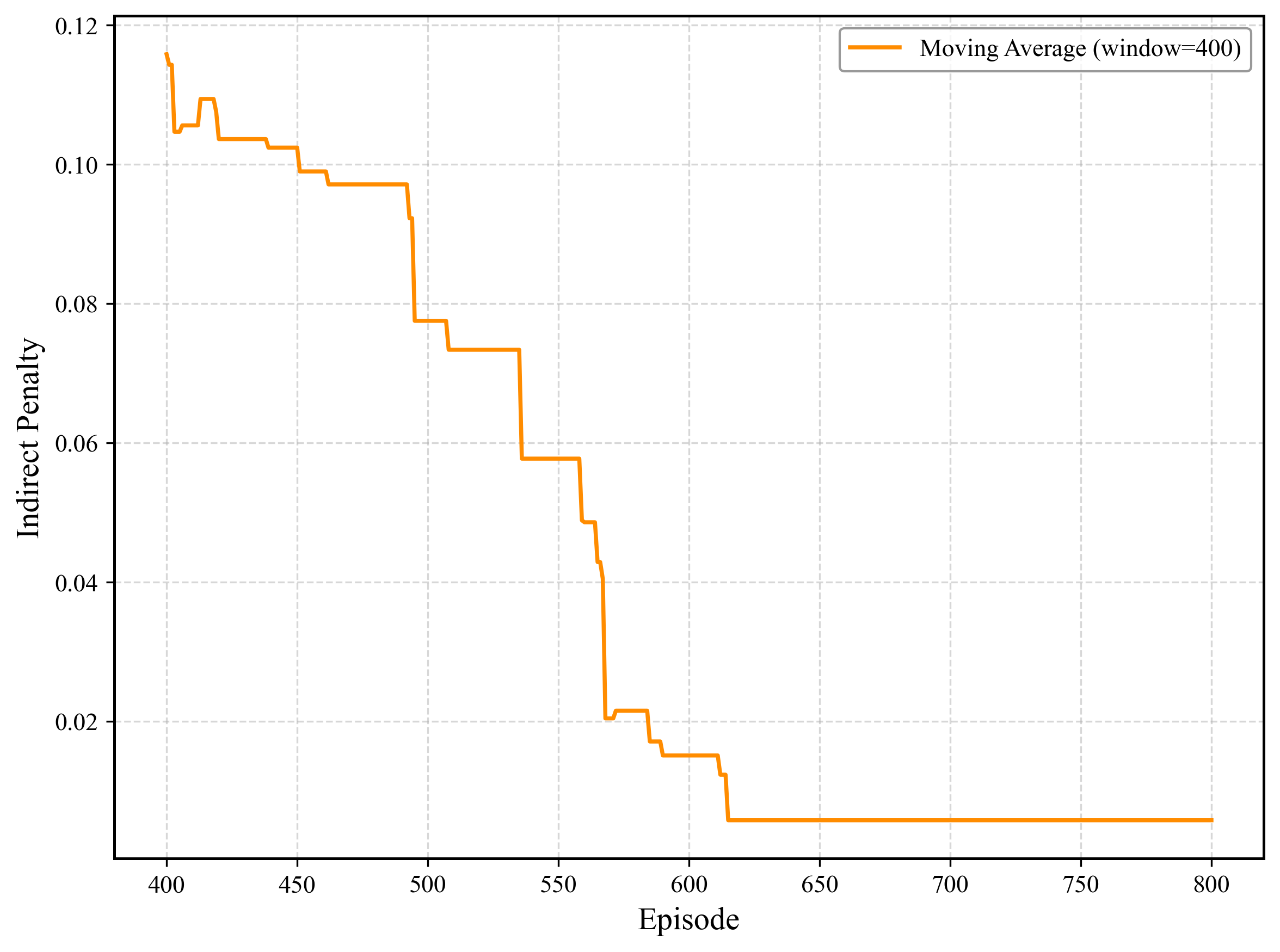}
        \captionsetup{font=footnotesize}
        \caption{Indirect penalty over episodes, showing steady reduction and stabilization, indicating minimized feature redundancy and convergence of the learning policy.}
        \label{fig:penalty}
    \end{minipage}
    \hspace{0pt}  
    \begin{minipage}[b]{0.48\textwidth}
        \centering
        \includegraphics[width=0.95\textwidth]{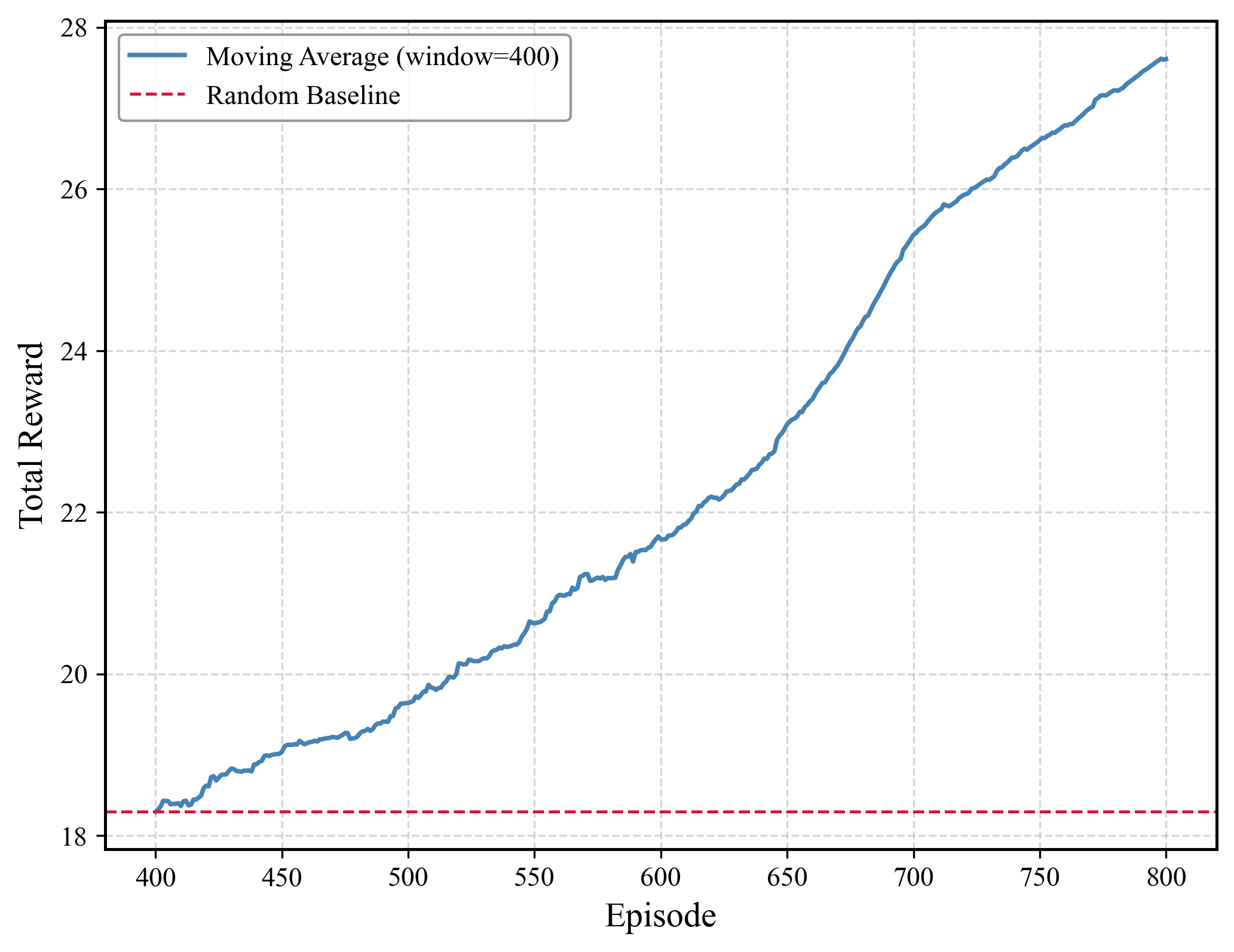}
        \captionsetup{font=footnotesize}
        \caption{Total rewards over episodes, the plot shows improved policy performance compared to the random baseline.}
        \label{fig:reward}
    \end{minipage}
    \label{fig:penalty_traj}
\end{figure}

We further evaluated the model’s performance with particular emphasis on the effectiveness of the multi-component reward function implemented in conjunction with the REINFORCE algorithm. This evaluation aimed to examine how the interaction between reward components—representing accuracy, fairness, and sparsity—guides the agent toward balanced policy optimization. Fig. \ref{fig:penalty} and Fig. \ref{fig:reward} illustrate the convergence of the learned policy during the feature selection process. This demonstrates the agent’s ability to progressively identify informative features while eliminating those that are irrelevant.

\begin{figure}[H]
    \centering
    \begin{minipage}[b]{0.48\textwidth}
        \centering
        \includegraphics[width=0.95\textwidth]{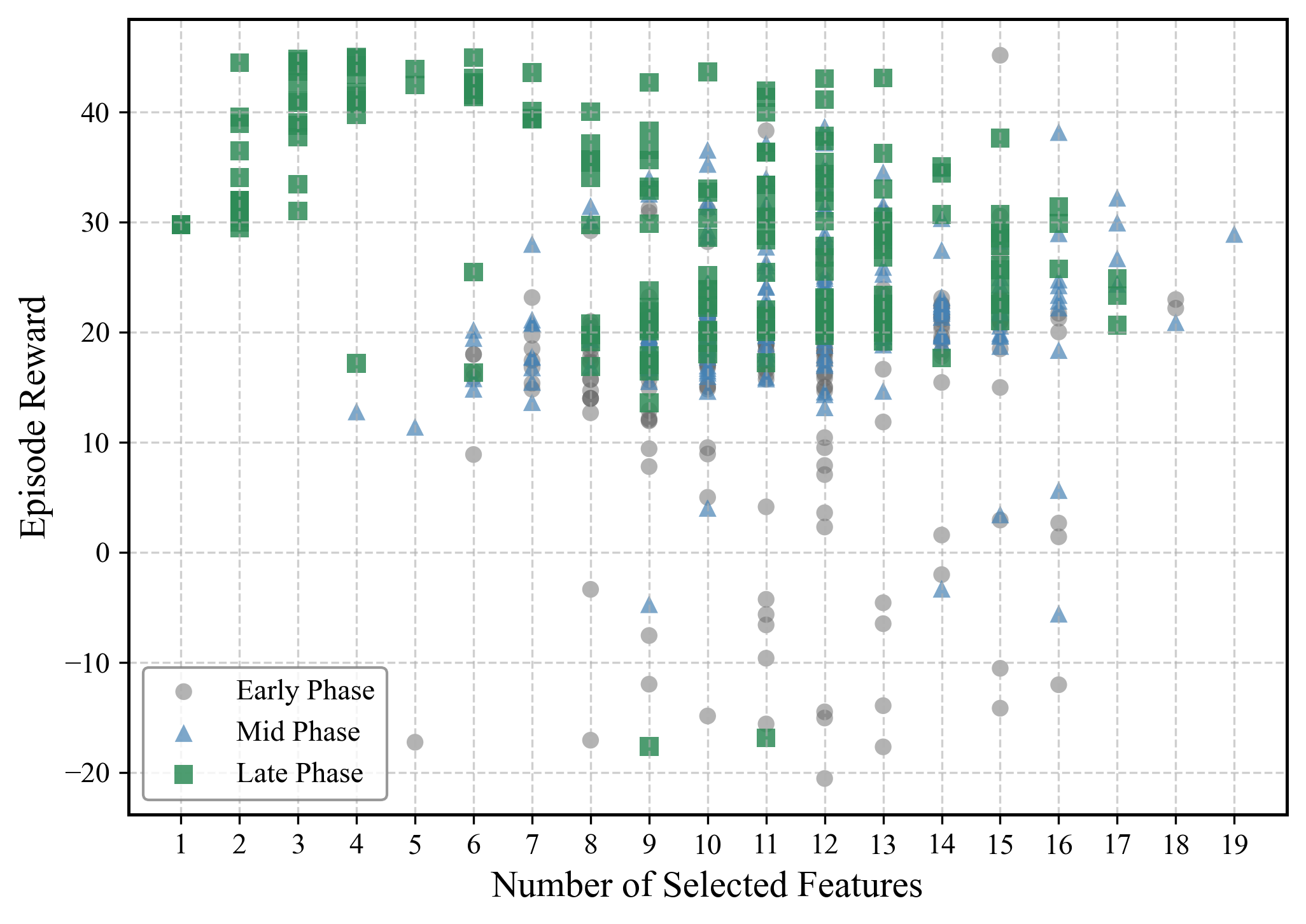}
        \captionsetup{font=footnotesize}
        \caption{Episode rewards versus number of selected features across training phases, shows policy improvement and convergence toward higher rewards in later training  stages.}
        \label{fig:policy_eval}
    \end{minipage}
    \hspace{0pt}  
    \begin{minipage}[b]{0.48\textwidth}
        \centering
        \includegraphics[width=0.95\textwidth]{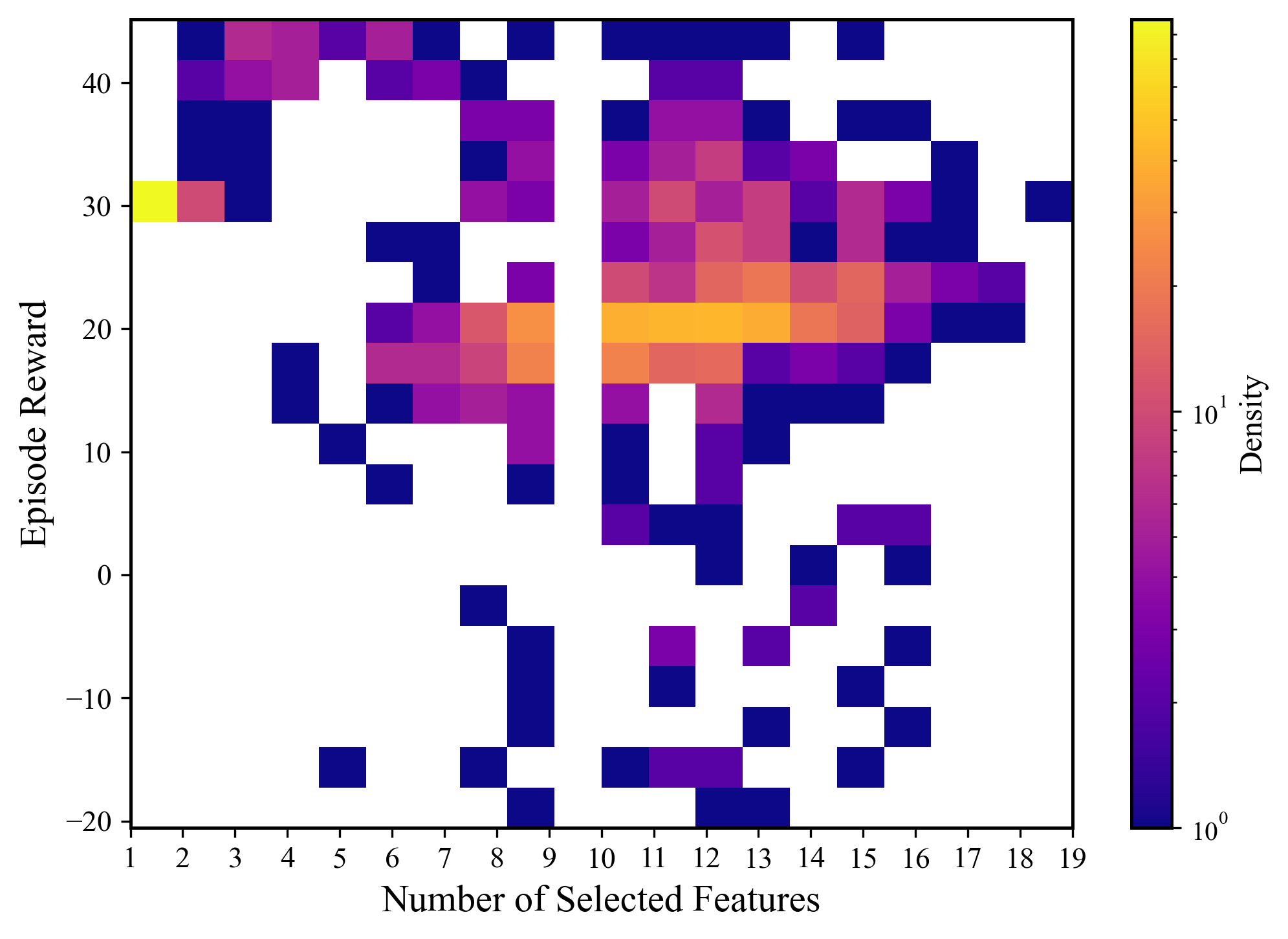}
        \captionsetup{font=footnotesize}
        \caption{Density heatmap illustrating reward distribution versus selected features, highlights optimal feature subsets concentrated between 9 and 13 features with consistent higher reward values.}
        \label{fig:heatmap}
    \end{minipage}
    \label{fig:reward_traj}
\end{figure}

In Fig. \ref{fig:penalty}, the plot shows the moving average indirect penalty based on training episodes. During the early training phase, the indirect penalty is quite high. This shows that the policy often picked less effective feature sets. As training continues, there is a steady downward trend. This trend reflects the agent's improved ability to minimize redundancy through policy improvement. After episode 600, the penalty drops sharply and stabilizes at a low value. This indicates that the learned policy successfully identifies non-redundant feature subsets that match the optimization goals.

Fig. \ref{fig:reward} illustrates the progression of RL agent performance in training episodes. The blue curve represents the moving average of the total reward showing a consistent upward trend. This indicates that the agent learns to select features more effectively as the training progresses. The red dashed line corresponds to the random baseline, serving as a reference for non-optimized selection. The steady divergence between the RL curve and the baseline highlights the agent’s improving policy and the effectiveness of the multi-component reward design in balancing accuracy, sparsity, and bias mitigation.

Fig.\ref{fig:policy_eval}, shows the distribution of episode rewards across the three training phases. During the early phase, the rewards exhibit large variance and several negative values, reflecting the agent’s initial exploration and unstable policy updates. As training progresses to the mid-phase, the agent begins to identify subsets of beneficial characteristics, indicated by a higher reward concentration between 10-14 selected characteristics. In the late phase, the reward values stabilize at consistently higher levels (approximately 30–45), signifying the convergence of the policy towards more optimal feature subsets. This progression illustrates how the policy gradient mechanism gradually refines its feature selection strategy through iterative policy updates. Fig.\ref{fig:heatmap}, presents a two-dimensional density heatmap. It provides a compact visualization of the frequency distribution between the selected number of features and the rewards of the episode. The regions of highest density, shown in yellow and orange, are concentrated around 9-13 selected characteristics with reward values between 20–30. This indicates that the policy consistently identifies this feature range as optimal under the designed multi-objective reward function. The heatmap further demonstrates that fewer than five or more than fifteen features rarely yield high rewards, emphasizing the trade-off between model complexity and performance inherent in the reward structure.

Our findings suggest that the proposed RL framework progressively learns to navigate the trade-offs among accuracy, sparsity, and fairness through continuous policy improvement. These results demonstrate that the agent achieves a stable balance between performance and fairness as learning progresses. However, its practical applicability within their respective domains can only be established by validation with the real data.

\section{Conclusion}
This study revel that RL can be effectively applied to feature selection in a manner that balances predictive relevance with fairness, thereby mitigating bias introduced through latent leakage. Our results show, as training progressed, the overall reward signal exhibited a steady upward trajectory, indicating stable learning dynamics. This improvement suggests that the agent effectively reconciled the tension between predictive accuracy and fairness through the multi-component reward formulation, progressively optimizing both objectives within an attainable trade-off boundary. The incorporation of correlation-based measures and directed acyclic graph based distance functions provided an effective way to approximate indirect causal pathways. However, this approach may be insufficient in contexts where the relationships among features are more complex or highly non-linear, limiting its general applicability.

Future research should aim to embed fairness constraints directly into the optimization procedure rather than distributing them across multiple reward components. Such integration could lead to more stable convergence and a better balance between accuracy and fairness at lower computational cost. Additionally, the adoption of advanced causal reasoning techniques may enable a more precise characterization of intricate interdependencies among features, thereby improving both fairness and robustness.

Overall, this work underscores the potential of RL as a tool for fairness-aware feature selection in high-dimensional settings. The purposed framework lays a foundation for further methodological innovations to reduce reliance on manual feature auditing and mitigate hidden bias introduced by proxy variables. Further research can extend this approach with enhanced causal inference methods and domain-specific validation to ensure its applicability in socially sensitive domains such as finance, healthcare, and criminal justice. In doing so, RL-based fairness-aware feature selection can move beyond theoretical contributions to provide actionable, equitable solutions in high-stakes decision-making systems.

\section*{Declarations}

\begin{itemize}
\item \textbf{Funding:} This work is not supported by any funding agency or grant.
\item \textbf{Conflict of interest/Competing interests:} None.
\item \textbf{Ethics approval:} Not applicable.
\item \textbf{Consent to participate:} Not applicable.
\item \textbf{Consent for publication:} Not applicable
\item \textbf{Availability of data and materials:} Applicable
\end{itemize}

\end{document}